\documentclass[letterpaper]{article}
\usepackage{uai2019}
\usepackage[margin=1in]{geometry}

\usepackage{times}

\usepackage{algorithm}
\usepackage[noend]{algpseudocode}
\usepackage{amsmath}
\usepackage{amssymb}
\usepackage{appendix}
\usepackage{subfig}
\usepackage{bbm}

\usepackage{adjustbox}
\usepackage{booktabs}
\usepackage{float}
\usepackage{subfig}
\usepackage{multirow}

\usepackage{graphicx}
\usepackage{amssymb}
\usepackage{epstopdf}
\usepackage{enumerate}
\usepackage{url}

\usepackage{graphics}
\usepackage{forest}
\usepackage{siunitx}

\usepackage{multirow}
\usepackage{mathrsfs}
\usepackage{times}
\usepackage{soul}
\usepackage{url}
\usepackage[utf8]{inputenc}
\usepackage{graphicx}
\usepackage{amsmath}
\usepackage{booktabs}
\usepackage{color, colortbl}
\usepackage{xcolor}
\newcolumntype{C}{>{\cellcolor[HTML]{003085}\color[HTML]{FFFFFF}\bfseries}c}
\newcolumntype{L}{>{\columncolor[HTML]{003085}[0pt][\tabcolsep]\color[HTML]{FFFFFF}\bfseries}l}

\usepackage{hyperref}
\hypersetup{
    colorlinks=true,
    linkcolor=blue,
    filecolor=magenta,      
    urlcolor = magenta,
    citecolor=blue
}

\newcommand{\jen}[1]{}
\renewcommand{\jen}[1]{{\color{blue} JN: {#1}}}

\newcommand{\mahak}[1]{}
\renewcommand{\mahak}[1]{{\color{red} MG: {#1}}}

\newcommand{\MethodName}[1]{}
\renewcommand{\MethodName}[1]{{DCPL {#1}}}

\newcommand{\specialcell}[2][c]{%
  \begin{tabular}[#1]{@{}c@{}}#2\end{tabular}}

\usepackage{footmisc}

\usepackage{color, colortbl}
\usepackage{xcolor}

\definecolor{Gray}{gray}{0.85}
\definecolor{LightCyan}{rgb}{0.88,1,1}
\definecolor{Green}{rgb}{0.8,1,0.8}
\definecolor{Yellow}{rgb}{1,1,0.8}
\definecolor{Orange}{rgb}{1,0.89,0.8}
\definecolor{Red}{rgb}{1,0.8,0.8}

\newcolumntype{a}{>{\columncolor{Gray}}c}
\newcolumntype{b}{>{\columncolor{LightCyan}}c}
\newcolumntype{g}{>{\columncolor{Green}}c}
\newcolumntype{d}{>{\columncolor{Yellow}}c}
\newcolumntype{e}{>{\columncolor{Orange}}c}
\newcolumntype{f}{>{\columncolor{Red}}c}

\usepackage{pifont}
\title{Cluster-Based Social Reinforcement Learning}

\author{ {\bf Mahak Goindani} \\
Department of Computer Science \\
Purdue University \\
West Lafayette, IN \\
mgoindan@purdue.edu \\
\And
{\bf Jennifer Neville}  \\
Departments of Computer Science and Statistics \\
Purdue University \\
West Lafayette, IN \\
neville@purdue.edu \\
}

\begin{document}

\maketitle

\begin{abstract}
\textit{Social Reinforcement Learning} methods, which model agents in large networks, 
are useful for fake news mitigation, personalized teaching/healthcare, and viral marketing, but it is challenging to incorporate 
inter-agent \textit{dependencies} into the models effectively due to network size and sparse interaction data.
Previous social RL approaches either ignore agents dependencies 
or model them in a computationally intensive manner. 
In this work, we 
incorporate agent dependencies efficiently in a compact model by 
clustering users (based on their \textit{payoff} and \textit{contribution} to the goal) and combine this with a method to easily derive personalized agent-level \textit{policies} from cluster-level policies. %
We also propose a dynamic clustering approach that captures changing user behavior. 
Experiments on real-world datasets 
illustrate that our proposed approach learns more accurate policy estimates and converges more quickly,  
compared to several baselines that do not use agent correlations or only use static clusters.


\end{abstract}

\section{INTRODUCTION} \label{intro}

Much of the existing work on multi-agent reinforcement learning (MARL) focuses on small number of agents. 
However, there are various applications that involve large number of interacting agents, for e.g., 
fleet management systems \cite{dejax1987survey}, urban transportation system consisting of a large number of vehicles, online advertising bidding agents \cite{wang2017display}, social networks with thousands of users who interact with each other (\cite{farajtabar2017fake, goindanisocial, upadhyay2018deep}). Thus, it is important to consider multi-agent systems with large number of interacting agents. 
Real-world social networks consist of thousands of users who interact with each other 
and are related in various ways (eg. \cite{farajtabar2017fake}). There is an opportunity to 
use the social network structure, which characterizes the relations and correlation between agents, to develop RL approaches that can scale for large number of agents. However, each user typically only interacts with a small number of other users in the network. This means the network interactions are overall sparse. 

We define \textit{Social Reinforcement Learning} as a sub-class of Multi-Agent Reinforcement Learning (MARL) for domains with large number of agents with relatively few (sparse) relations and interactions between them.
%
%
The objective is to learn a \textit{policy} $\pi$ that maps the network \textit{state} (over $N$ users) to \textit{actions} (over $N$ users). 
In a real-world social network, users
interact with each other, which  
leads to \textit{dependencies} between their \textit{actions} (due to peer-influence). For example, if one user tweets more, that may influence their followers to tweet more as well.
It is important to capture inter-agent dependencies
to learn accurate policies for social RL systems \cite{goindanisocial2020_workshop}. Thus, to learn the policy for a single user, we need to consider the actions of all $N$ users, resulting in at least $N^2$ parameters to learn $N$ user policies with $N$ actions per policy. 
However, at the same time social networks are typically \textit{sparse} where agents only interact with a constant number of other agents, i.e., $O(N)$ interactions, as opposed to dense networks where all agents interact, i.e., $\Omega(N^2)$ interactions. 
Therefore, learning accurate policies, given $O(N)$ (sparse) interactions, is difficult since there are insufficient observations to capture ${\sim}N^2$ agent dependencies. 
This is, particularly, more challenging for large $N$ due to the curse of dimensionality, resulting in increased variance. 

Traditional MARL approaches do not scale for large numbers of agents. Learning a separate model for each user independently \cite{He2016OpponentMI, Battaglia2016InteractionNF, Hausknecht2016DeepRL, Yliniemi2016MultiobjectiveMC, Dusparic2009DistributedWM, cao2018emergent, leibo2017multi, Hughes2018InequityAI, jin2018real, mason2016applying, Kim2019MessageDropoutAE, Jaques2018IntrinsicSM, zhang2018fully, shu2018m, sukhbaatar2017intrinsic, Lazaridou2017MultiAgentCA, Mannion2017MultiagentCA, raileanu2018modeling, omidshafiei2017deep, Palmer2018NegativeUI, mannion2015learning, Kumar2017FederatedCW, yang2018mean_MFRL, sen2007emergence}
without considering agent dependencies results in less accurate policy estimates, and in addition, requires learning $N$ complex models that is infeasible for large $N$. 
Some recent work employed joint policy learning \cite{farajtabar2017fake,goindanisocial,sukhbaatar2016learning_commnet} to capture agent dependencies. However, these do not address the problem of sparse user interactions, and still estimate ${\sim}N^2$ parameters resulting in high computational cost. 
Also, due to the high-dimensionality of joint state/actions representation in these methods, 
the impact of an individual's state features in learning her own policy diminishes with increase in $N$, resulting in noisy estimates as the policy parameters overfit to a single type of users who may be in majority.
%
In addition, it makes it infeasible to model the reciprocity in agent interactions, because $N^2$ agent \textit{pairs} would lead to at least  $N \!\times\! N^2\!=\!N^3$ parameters. 
To avoid this, the above approaches considered only 
individual (i.e. $N$) actions per user.

To address these challenges, we propose a \textit{Dynamic Cluster-based Policy Learning} approach \MethodName that utilizes the properties of the social network structure and agent correlations to obtain a compact model to represent the system dynamics.
%
Specifically, we propose to cluster similar users in order to reduce the effective number of policies to be learned, and overcome the problem of sparse data by aggregating the interactions of similar users. We then develop a method to easily derive personalized agent-level actions from cluster-level policies by exploiting 
variability in agents' behavior. 
Thus, we reduce the problem of learning policies for $N$ users to that for $\mathcal{C}$ clusters $(\mathcal{C} \ll N)$,
and hence, the model dimensionality 
from ${\sim}N^3$ (with pairwise agent interactions) to ${\sim}\mathcal{C}^3$, 
to allow efficient and effective \textit{joint} policy learning considering all users. 

Learning policies from the global system reward alone, without considering individual contributions, can result in noisy estimates. 
To address this, we design clustering features, motivated by \textit{Difference Reward} \cite{devlin2014potential}. 
%
Specifically, we define \textit{contribution} features to measure a user's efficacy given other agents' actions, and, \textit{payoff} features to determine how responsive an agent is to the policy applied in the past. 
We use these features (via clusters) 
to learn agents' effectiveness early on, for better exploring the action space, without increasing the state space, in order to
speed-up convergence. 
%
%
Moreover, agent interactions in a social network are dynamic, and thus, we update the policy after regular time-intervals.
Since we want clusters to reflect the effects of applying the updated
policy, we need to dynamically update cluster memberships while
learning the policy. However to learn across time steps, it is important to ensure that the clusters
are aligned---to achieve this, we propose a
\textit{weighted centroid} clustering approach. 
To our best knowledge, our proposed approach is the first to consider policy learning while dynamically clustering users in MARL for social networks.


We evaluate performance compared to different static clustering and non-clustering policy learning methods, using real-world Twitter datasets.
Results show that compared to other baselines, our dynamic cluster-based approach 
is able to 
learn better \textit{resource allocation} and  
achieve higher \textit{reward}, 
by learning users' effectiveness via \textit{payoff} and \textit{contribution} features.
Clustering enables faster convergence of policy learning, by reducing the effective number of policies and producing a compact model with lower dimensionality.
In addition, our approach has lower variance, which indicates that clustering is able to overcome sparsity in agent interactions, and learn personalized policies that are not biased towards the majority type of users. 

\section{RELATED WORK} \label{related_work}

\begin{table*}[!t]
\centering
\caption{Comparison of Social RL approaches (highlighted) with existing MARL approaches. $`-'$ represents that a given attribute is not defined for the problem setting considered in the corresponding approach.}
{
\begin{tabular}{|c|c|c|c|c|c|}
    \toprule
    \multirow{4}{*}[-0.5\dimexpr \aboverulesep + \belowrulesep + \cmidrulewidth]{\textbf{Approach}} & \multicolumn{3}{|c|}{\textbf{Data/Environment}} & \multicolumn{2}{|c|}{\textbf{Model}} \\
    \cmidrule(l){2-6}
    {} & \textbf{\specialcell{Network \\Density}} & \textbf{\specialcell{Link \\Types ($d$)}} & \textbf{Space}  & \textbf{\specialcell{Maximum \\Dimensionality}} & \textbf{\specialcell{Effective \\ Dimensionality}}  \\
    \midrule
    \rowcolor{Yellow}  
    \cite{farajtabar2017fake} & $O(N)$ & 1 & C  & ${\sim}N^2$ & ${\sim}N^2$  \\
    \rowcolor{Yellow} 
    \cite{upadhyay2018deep} & $O(N)$ & 2 & C  & ${\sim}N$ & ${\sim}N$  \\
    \rowcolor{Yellow} 
    \cite{goindanisocial} & $O(N)$ & 2 & C  & ${\sim}N^2$ & ${\sim}N^2$  \\
    \rowcolor{Yellow} 
    \cite{sukhbaatar2016learning_commnet} & $\Omega(N^2)$ & 1 & D  & ${\sim}N^2$ & ${\sim}N^2$ \\ 
    \cite{yang2018mean_MFRL} & $\Omega(N^2)$ & 1 & D  & ${\sim}N^2$ & ${\sim}N$ \\
    \cite{nguyen2017collective, Nguyen2018CreditAF} & $-$ & 1 & D  & ${\sim}N^2$ & ${\sim}N^2$  \\
    \cite{Lin2018EfficientLF} & $-$ & 1 & D  & ${\sim}N^2$ & ${\sim}k$ \\
    \cite{Leibo2018MalthusianRL} & $-$ & 1 & D  & ${\sim}kN$ & ${\sim}kN$ \\
    \bottomrule
\end{tabular}
}
\label{related_work_comparison_table}
\end{table*}

The number of agents in Social RL problems are large as opposed to much of previous work in MARL that considers small $N$
\cite{malialis2015distributed, mason2016applying, Prolat2017AMR, Hausknecht2016DeepRL, Tampuu2017MultiagentCA, Liu2018EmergentCT, Taylor2014AcceleratingLI, He2016OpponentMI}.
Although, there are a few problems that involve large $N$  \cite{sukhbaatar2016learning_commnet, Nguyen2018CreditAF, nguyen2017collective, Lin2018EfficientLF, Leibo2018MalthusianRL},
they consider dense agent interactions 
that allows to capture 
agent dependencies more easily (e.g. \cite{yang2018mean_MFRL}). On the contrary, 
Social RL problems have sparse 
interaction data \cite{farajtabar2017fake,goindanisocial,upadhyay2018deep} that makes it challenging to capture these dependencies for learning accurate policies \cite{goindanisocial2020_workshop}.

We present a comparison of Social RL with state-of-the-art MARL approaches, based on different environment factors and model settings in Table \ref{related_work_comparison_table}. 
The existing approaches that consider a small number of agents differed only on a few aspects. Thus, we focus on only those approaches that have considered a large number of agents. Learning policies for 
Continuous (C) spaces is more challenging than that for
Discrete (D) spaces, 
and much of the RL models are only applicable to discrete spaces 
(e.g., \cite{Liu2018EmergentCT}). 
However, Social RL problems generally have continuous spaces describing 
complex network activities. 
It is important to study the computational complexity (number of parameters required to learn policies for all agents) of different approaches. Based on the agent dependencies considered, each model has a maximum number of parameters required for learning the policy for all agents, referred to as the \textit{Maximum Dimensionality} in Table \ref{related_work_comparison_table}. 
\textit{Effective Dimensionality} corresponds to the effective (actual) number of parameters learned based on the approximations/assumptions made by the model. 
For example, \cite{Lin2018EfficientLF} considers an aggregate of agents to
obtain a smaller effective number of agents. 
They learn policies using parameter sharing, resulting in effective number of parameters as ${\sim}k$. 
However, their approach is designed for small restricted discrete action spaces \cite{nguyen2018deep}.
%
\cite{nguyen2017collective,Nguyen2018CreditAF} 
use an aggregate statistic of agents' actions along with certain strong assumptions on the state-transition model, to capture ${\sim}N^2$ dependencies. 
%
%
However, Social RL problems usually have continuous spaces describing 
the intricate user interactions and network activities, and the complex state-transition dynamics are generally unknown.

There are several real-world applications involving social networks that can be modeled using Social RL. Social RL helps to estimate user responses (likes, comments, shares) considering effects of peer-influence \cite{goindanisocial}. It can be used to understand customer demands, and learn strategies to incentivize competing products by capturing changing user interests and feedback, e.g., recommender systems with thousands of users and very few user-user/user-item interactions such as news recommendation \cite{zheng2018drn},  recommending potential collaborators \cite{zhang2017dynamic}.
Also, it is important to learn credit assignment when multiple agents share a limited resource. Social RL can help identify \textit{influential} users for viral marketing or resource allocation under fixed budget. 
%
Social RL has been used to find solutions for critical problems involving social networks such as fake news mitigation and learning optimal intensities for news diffusion.
Thus, there is a need to develop efficient solutions for Social RL problems that can scale for large number of agents, and overcome the problem of sparse interaction data. 

\cite{farajtabar2017fake} proposed to 
mitigate fake news by spreading more true news, 
on Twitter that is a sparse network with thousands of users. 
They integrated user activities and interactions (via tweets), characterized using Multivariate Hawkes Process (MHP) in a RL framework, and learned interventions for the MHP via policy optimization in a Markov Decision Process (MDP). 
\cite{goindanisocial} extended their approach and proposed to model another form of user interactions, i.e. feedback, quantified as the number of likes received on sharing a post. The feedback is integrated in a Social RL framework that helps 
appropriate selection of users and efficient allocation of incentives among them under budget constraints.
\cite{upadhyay2018deep} considers 
agent interactions via tweet and feedback for learning optimal strategies in personalized teaching and viral marketing domains, to increase the recall probability among learners, and \textit{visibility} among followers on Twitter, respectively.
\cite{farajtabar2017fake,goindanisocial} developed joint learning approaches for fake news mitigation on Twitter, by modeling agent interactions via tweets and \textit{likes}. 
They consider individual (i.e. $N$) actions and estimate ${\sim}N^2$ parameters resulting in high computational cost for large networks. Also, they do not address the problem of sparse interactions that leads to increased variance in estimates. 
In contrast, 
we reduce the model dimensionality by clustering similar users. 
This allows us to efficiently learn a more compact model and facilitates modeling actions per user {\em pair} (i.e. $N^2$ actions) to capture reciprocity. 
Moreover, it addresses the problem of sparse interaction data by aggregating the interactions of similar users.
\cite{upadhyay2018deep} 
learned a separate model for each agent. However, this ignores inter-agent dependencies, and is infeasible for large $N$.

\cite{goindanisocial} did not distinguish between tweets and retweets,  
and learned the same action for both activities.
However, the diffusion patterns of tweets and retweets have large differences, especially, for fake news \cite{carchiolo2018terrorism}.
%
Moreover, retweet network is very different from followers network on Twitter with a low level of reciprocity \cite{vosoughi2018spread}.
%
To capture these differences, we consider tweet and retweet as separate network activities,
resulting in three different types of agent interactions. 
Also, we learn actions per user pair to capture the reciprocity in user interactions, i.e., $N^2$ actions. 
However, this leads to an increase in the model dimensionality, 
particularly 
for joint learning,
resulting in large number of parameters (${\sim}N^3$), and high computational cost. 

To overcome this, and learn better resource allocation strategies, we propose to first cluster users with latent features (based on their past behavior and contribution to the objective).
%
%
Previous work utilized \textit{symmetry} between states and/or agents to reduce the size of the Markov Decision Process \cite{zinkevich2001symmetry, ravindran2002model, mahajan2017symmetry}.
%
Our clustering approach is based on the observation that 
similar users tend to show similar behavior, and thus we propose to learn similar policies for them.
This has also been utilized to cluster users in mobile health domain
\cite{zhu2018group,el2018personalization}. 
%
%
However, they did not consider user interactions and dependencies, clustered users only once, and assumed the cluster assignments remain fixed (\textit{static}) throughout policy learning. 
%
In our setting, we consider 
dynamic user interactions, and the policy is updated after regular time-intervals. 
Thus, there is a need to dynamically update cluster assignments, and we propose an approach 
to ensure cluster alignment at different time-steps.


Relational reinforcement learning (RRL) combines RL with relational learning or inductive logic programming \cite{muggleton1994inductive} to represent states, actions, and policies using the structures and relations that identify them \cite{ponsen2010learning}. The structural representations facilitate 
solving problems at an abstract level, and thus RRL approaches provide better generalization.
%
RRL has been used in multi-agent systems to share information among agents, to learn better individual policies \cite{croonenborghs2005multi}. 
%
Previous work used first-order representations to achieve effective state/action abstractions \cite{dvzeroski2001relational,asgharbeygi2006relational,ponsen2010learning}.  
Some recent work has also used relational interactions between agents 
for policy learning in multi-agent systems
\cite{grover2018evaluating,grover2018learning,tacchetti2018relational,zambaldi2018relational}.
While relational languages facilitate compact state space representations, in multi-agent scenarios, relational interactions further increase the size of the state space and increase reasoning complexity~\cite{croonenborghs2005multi}.
Thus, these approaches are limited to discrete spaces with small number of agents that have dense interactions between them.
We take motivation from relational learning to capture agent relations and interactions, however, we do not directly use 
RRL due to the limitations described above.

\section{SOCIAL REINFORCEMENT LEARNING}

\subsection{PROBLEM DEFINITION} \label{prob_def}

We define \textit{Social Reinforcement Learning} as a sub-class of Multi-Agent RL, that considers large number of agents with interactions between them. 
Specifically, we consider a social network setting with $N$ users. Each user $i \in \{1,...,N\}$ is an agent.  
Let $G = (V, E)$ represent the social network graph, where each node $v_i \in V$ corresponds to user $i$, and edge $E_{ij} = 1$ if there exists an edge between agent $i$ and $j$, and 0 otherwise.
In addition, users perform $d$ different activities (e.g., tweet, comment, like) in the social network. We say that the network is dense when all the agents interact, i.e., the  number of interactions is $\Omega(N^2)$, and the network is sparse when agents only interact with a constant number of other agents, i.e., $O(N)$ interactions. 

Let $\textbf{s}_i \in \mathbb{R}^{d} \; (\textbf{s}_i \geq 0)$ be the state of user $i$. This corresponds to the $d$ activities the user performs. Then the state of the network represents the activities over the $N$ users, $\textbf{s} = \{\textbf{s}_i\}_{i=1}^N$. The network activities are dynamic in nature, and thus, we have different states of the network at different time-stamps. Let $\textbf{s}_t$ represent the network state at time $t$. We consider a finite horizon setting, i.e., we observe the activities up to time $T$.
An action $a_{d, i} \in \mathbb{R}$ corresponds to a modification to the $d$-th activity of user $i$. Let $\textbf{a}=\{a_{d, i}\}_{i=1}^N$ refer to a set of actions, one for each user in the network. 
Collective actions of all users lead to a change in the state of each user, and consequently, a change in the overall network state. Let $\mathcal{T}(\textbf{s}, \textbf{a}, \textbf{s}^\prime)$ be the probability of transitioning to the network state $\textbf{s}^\prime$ after performing actions $\textbf{a}$ in network state $\textbf{s}$. 
While there may be a reward for each agent $i$ and/or each activity $d$, for generality we will consider a network-based (i.e., common) reward $R(\textbf{s},\textbf{a}) \in \mathbb{R}$ that is based on the network state and collective actions of users. 
The goal is to learn a policy that maps the network state at time $t$ to  actions for each user, i.e., $\pi: \textbf{s}_t \rightarrow \textbf{a}_t$ such that the total expected discounted reward $\sum\limits_{t=1}^T \gamma^{t-1} \mathbb{E}[R_{t}]$ is maximized, where $\gamma \in [0, 1)$ is the discount rate. 

\begin{figure}[!t]
    \centering
    \includegraphics[height=5cm]{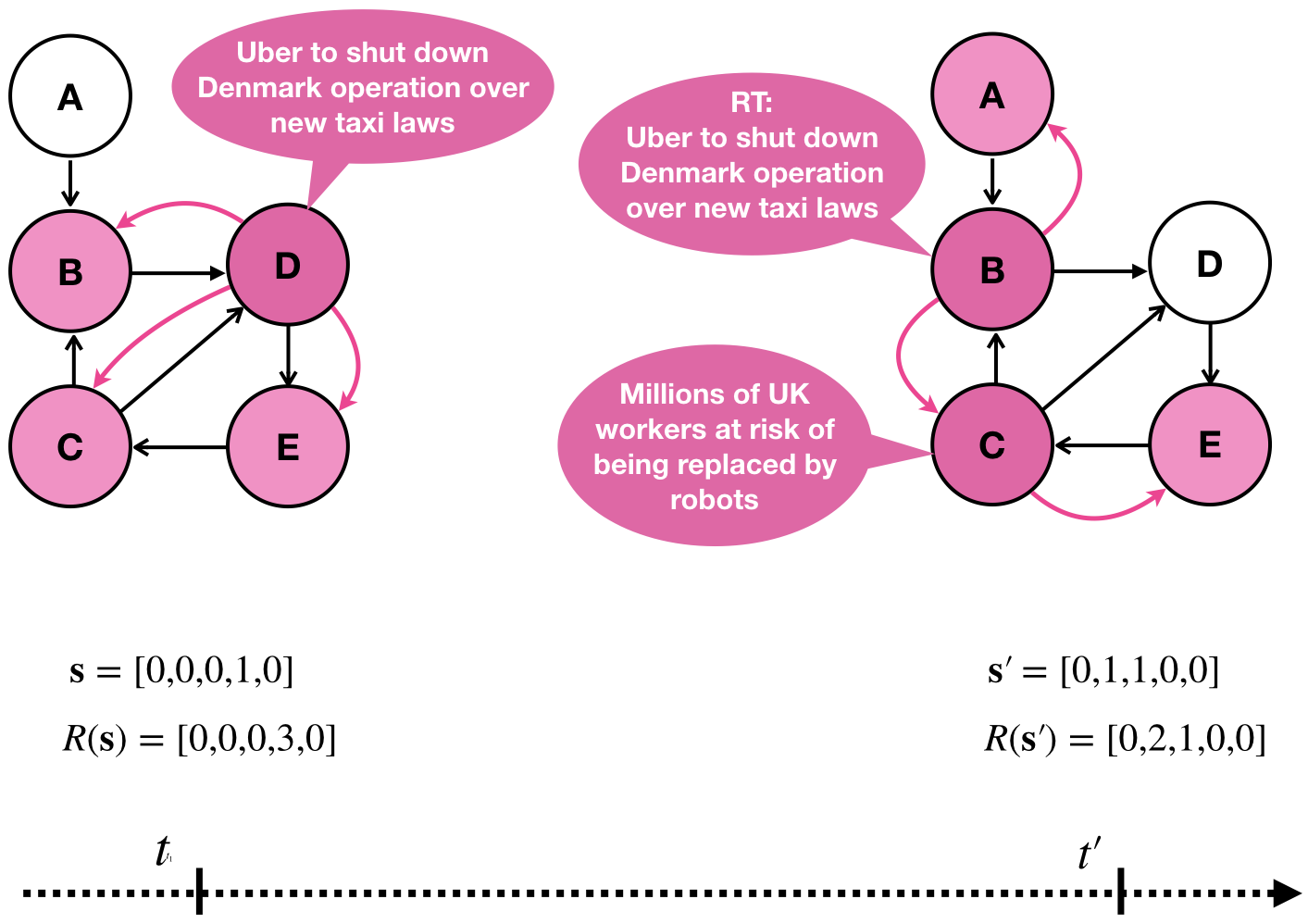}
    \caption{\small{Social network with agent interactions. Colored nodes represent active users (darker shade for users who are tweeting, lighter shade for people exposed to tweets). State is given by the number of tweets for users [A,B,C,D,E], and reward is calculated as the number of users exposed. Users' collective actions of tweeting news lead to a transition from state $\textbf{s}$ to $\textbf{s}^\prime$.}}
    \label{network_fig_workshop}

\end{figure}

Because users interact in the network, actions taken for one user may impact the state (i.e., activity) of other users. More specifically due to peer-influence, changes in the activity of user $i$ may influence the likelihood of user $j$'s activities in the future if $i$ and $j$ are connected in $G$, either directly (i.e., $E_{ij}=1$) or through a longer path. For example, if one user tweets more, that may influence their followers to tweet more as well. Fig. \ref{network_fig_workshop} shows a network of users who interact via tweeting news. 
Thus, agent interactions lead to dependencies among agent actions and state transitions. 
Specifically, the state transition distribution does not depend on the individual user activities $\textbf{s}$ and actions $\textbf{a}$ alone, but also depends on the network $G$,
i.e., $P(\textbf{s}^\prime | \textbf{s}, \textbf{a}, G) \neq P(\textbf{s}^\prime | \textbf{s}, \textbf{a})$.  Also note that the underlying state transition distribution is not always explicitly known as it depends on the network dynamics, which is the model-free RL setting. Since the network reward $R$ depends on the state transition distribution $\boldsymbol{\mathcal{T}}$, and  user $i$'s activity can be moderated both by actions taken with respect to $i$ and actions taken for other users $j\!\neq\! i$, it is important to consider the effect of network structure when learning the policy $\pi$. 
At the same time, social networks are typically sparse with only $O(N)$ interactions between agents. Thus, the challenge is to capture ${\sim}N^2$ agent dependencies 
given these sparse interactions. 

\subsection{CHALLENGES}

We present different challenges for learning Social RL models and discuss potential opportunities to address them using properties of the network structure and agent correlations. 

\subsubsection{High-Dimensionality}

The joint action space grows larger with $N$, 
as it considers the actions of all agents for learning the policy 
i.e., $O(N^2)$ dependencies, resulting in high computational complexity.
Also, the impact of an individual's state features in learning her own policy diminishes with increase in $N$, which can result in noisy estimates as the policy parameters overfit to a single type of users who may be in majority.
Additionally, independent learning requires training $N$ models. This is impractical for thousands of agents, especially, when the policy function approximators are complex (e.g. Deep Neural Networks).

\subsubsection{Sparsity}

Typically, social networks are sparse with only $O(N)$ agent interactions. Thus, learning accurate policies becomes difficult as there are not sufficient samples available to capture $O(N^2)$ agent dependencies. 
This is particularly more challenging for large $N$ due to curse of dimensionality, resulting in large variance in estimates.


Next, we present potential solutions to address the above challenges using properties of the network structure and agent correlations. Our key insight is to utilize the correlations between agents to reduce the size of the Markov Decision Process (MDP). Specifically, we propose to cluster similar users in order to reduce the effective number of policies to be learnt, and overcome the problem of sparse data by aggregating the interactions of similar users. Thus, we can obtain more accurate policy estimates along with faster convergence due to large reduction in dimensionality and the variance.
We demonstrate the utility of our proposed approach using the application of  
Fake News Mitigation in social networks as follows. 

\section{CLUSTER-BASED SOCIAL RL APPROACH} \label{approach}

\subsection{APPLICATION: FAKE NEWS MITIGATION} 

We consider a social network with $N$ users who interact via $d$ different
network activities (e.g. tweet, like).
Each user $i \in \{1,...,N\}$ is an agent.  
Let $\textbf{G}$ 
represent the followers adjacency matrix for the social network graph, where $G_{ij} = 1$ if $j$ follows $i$, and 0 otherwise. 
%
%
Our data contains a temporal stream of events with the time horizon [0, $\mathcal{T}$) divided into $K$ stages, each of time-interval $\Delta T$, where stage $k \in [1, K]$ corresponds to the time-interval $[\tau_k, \tau_{k+1})$. 
We consider three types of events characterizing user activities, corresponding to tweets ($\mathscr{T}$), retweets ($\mathscr{R}$), and likes ($\mathscr{L}$). 
We represent the tweet or retweet events using $e=(t, i, h, z)$ where $t$ is the time-stamp at which user $i$ shares a post of type $z$ = $\mathscr{T}$ or $\mathscr{R}$, with label $h$ = $F$ (Fake) or $T$ (True). 
(Note that our goal is not to detect, but to mitigate the impact 
of fake news. Thus, we consider the (re)tweets labeled fake/true apriori.)
Like events are represented as $l(u, i, t)$ indicating user $i$ likes user $u$'s post at time $t$. 
%
To enhance readability, we provide a list of symbols used in the paper along with their description in Table \ref{notations}. 

\begin{table*}[!h]
\centering
\caption{Notations}
\scalebox{0.85}
{
\begin{tabular}{lll}
    \toprule
    {} & \textbf{Symbol} & \textbf{Description} \\
    \midrule
    \multirow{6}{*}[-0.5\dimexpr \aboverulesep + \belowrulesep + \cmidrulewidth]{{\specialcell{\textbf{Data} \\ \textbf{Description}}}} & $N$ & Number of Agents (users) \\
    & $\textbf{G}$ & Social Network \\
    & $h$ & Label for news: Fake (F) or True (T) \\
    & $z$ & Type of network activity: Tweet ($z=\mathscr{T}$) or Retweet ($z=\mathscr{R}$) or Like ($z=\mathscr{L}$) \\
    & $\mathcal{T}$ & Time-Horizon \\
    & $K$ & Number of Stages \\
    & $k \in [1,K]$ & $k-$the Stage corresponds to Time-Interval $[\tau_k, \tau_{k+1})$ \\
    & $\Delta T$ & Time-interval Length, i.e., $\tau_{k+1} - \tau_k$ \\
    \midrule
    \multirow{10}{*}[-0.5\dimexpr \aboverulesep + \belowrulesep + \cmidrulewidth]{{\specialcell{\textbf{Activity} \\ \textbf{Processes}}}} & $\lambda_{h, z, i}$& Intensity function for user $i$ to share a post with label $h$ = F or T, and type $z$ = $\mathscr{T}$ or $\mathscr{R}$ \\
    & $\mu_{h, z, i}$& \specialcell{Base Exogenous Intensity for user $i$ to share a post with label $h$ = F or T, type $z$ = $\mathscr{T}$ or $\mathscr{R}$, \\ whose intensity is governed by Multivariate Hawkes Process (MHP)} \\
    & $\boldsymbol{\Phi}_z$& Kernel Adjacency Matrix for MHP for network activity of type $z$\\
    & $\omega_{h, z}$& Kernel Decay Parameter for Exponential Hawkes Kernel \\
     & 
    $\mathcal{N}_i(t,h,z)$ & Number of posts of label $h$ (= T or F), and type $z$ (= $\mathscr{T}$ or $\mathscr{R}$), made by user $i$ upto time $t$ \\
    & ${n}_{k, i}(h,z)$ & Number of posts of label $h$ (= T or F), and type $z$ (= $\mathscr{T}$ or $\mathscr{R}$), made by user $i$ in stage $k-1$ \\ 
    & ${W}_i(t)$ & Number of likes received by user $i$ upto time $t$ \\
    & ${w}_{k,i}$ & Number of likes received by user $i$ in stage $k-1$ \\ 
    \midrule
    \multirow{20}{*}[-0.5\dimexpr \aboverulesep + \belowrulesep + \cmidrulewidth]{{\specialcell{\textbf{Cluster-based} \\ \textbf{Policy} \\ \textbf{Learning}}}} & 
    $\mathcal{C}$& Number of Clusters \\
    & $\textbf{s}_{U,k, i}$ & State of User $i$ at stage $k$\\
    & $\textbf{s}_{U,k}$ & User-level Network State at stage $k$\\
    & $\textbf{s}_{C,k, m}$ & State of Cluster $m$ at stage $k$\\
    & $\textbf{s}_{C,k}$ & Cluster-level Network State at stage $k$\\
    & $\textbf{a}_{U, z, k, i}$ & Action of User $i$ at stage $k$ for activity $z$ = $\mathscr{T}$ or $\mathscr{R}$\\
    & $\textbf{a}_{C, z, k, m}$ & Action of Cluster $m$ at stage $k$ for activity $z$ = $\mathscr{T}$ or $\mathscr{R}$\\
    & $\boldsymbol{\mathcal{M}}_k \in \{0, 1\}^{N \times \mathcal{C}}$ & Cluster Membership Matrix at stage $k$\\
    & $\textbf{X}_{k,i}$ & Clustering Features for user $i$ at the start of stage $k$ \\
    & $\beta_{z, k, i}$ & \textit{Payoff} Feature for user $i$ for stage $k$, corresponding to activity of type  $z$ = $\mathscr{T}$ or $\mathscr{R}$ \\
    & $\Omega_{d, k,i}$ & \textit{Contribution} Feature for user $i$ for stage $k$, corresponding to activity of type  $z$ = $\mathscr{T}$ or $\mathscr{R}$ \\
    & $\textbf{Y}_{k, m}$ & Centroid of cluster $m$ at stage $k$ \\
    & $\pi_{C,m}$ & Policy for Cluster $m$\\
    & $\pi_{U}$ & Function to obtain User-level Actions from the Cluster-level Actions (Algorithm 2, Main Paper)\\
    & $d$ & Number of activities in the network\\
    & ${R}$& Reward \\
    & $V(\textbf{s}_{U,k})$& Value-function for network state $\textbf{s}_{U,k}$  \\
    & $\gamma$& Discount rate governing the relative importance of immediate and future rewards \\
    \bottomrule 
\end{tabular}
}
\label{notations}
\end{table*}

A quantitative measure of the impact of fake and true news is the number of people exposed. 
Let $\mathcal{N}_i(t, h, z)$ represent the number of times user $i$ shares news of type z = $\mathscr{T}$ or $\mathscr{R}$, with label $h$ = $F$ or $T$, up to time $t$. 
Then the number of times a user $i$ is exposed to news up to time $t$ corresponds to $G_{ji} \cdot \mathcal{N}_j(t, h, z)$ ($j \in [1,N]$). 
Let $W_{i}(t)$ be the number of likes received by user $i$ upto time t.
%
Our goal is to incentivize users to share true news---to ensure that users receive at least as much true as fake news. 
Algorithmically, we want to increase the probability of sharing true news in a targeted fashion 
by learning an efficient strategy to 
allocate incentive among users, based on properties of the network $\textbf{G}$ and the effects of past user interactions.  
%
Specifically, given the network {\em state} $\textbf{s} \in \mathbb{R}^{dN}$ corresponding to $d$ user activities  
(described further in Sec. \ref{state_features}), 
we want to learn an incentivization policy $\pi: \textbf{s} \rightarrow \textbf{a}$ to obtain incentive {\em actions} $\textbf{a} \in \mathbb{R}^{N + N^2} (\textbf{a} \geq 0)$, where tweet actions are per user (i.e., $N$) and retweet actions are per user pair (i.e., $N^2$).
Incentive corresponds to increasing the likelihood of users sharing true news, 
and can be realized (in real-world) 
by creating certain accounts whose sharing rate is increased according to the learnt policy or by using external motivation (eg. money). 
%
Note that fake news mitigation is a multifaceted problem, and the reward function depends on the different applications. 
Thus, in contrast to the simple reward described in Sec. \ref{prob_def}, now we will consider \textit{reward} as the correlation between exposures to fake and true news. This is based on the idea that users exposed more to true news must be exposed more to fake news \cite{farajtabar2017fake}. 

%

\begin{figure*}[!t]
\centering 
    {\includegraphics[width = 16.5cm]{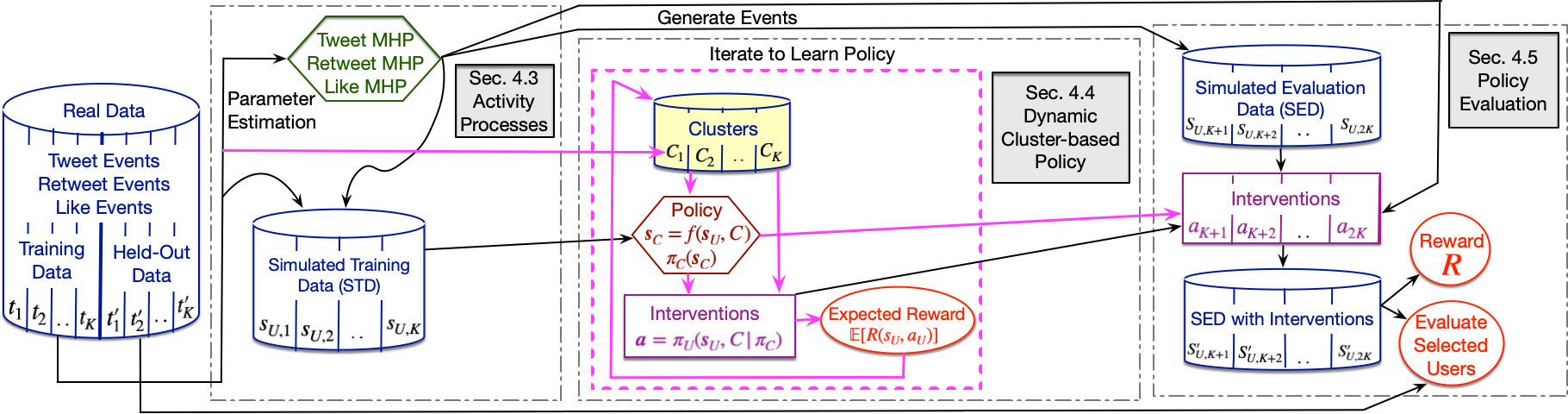}}
    \caption{Overview of Dynamic Cluster-Based Policy Learning and Evaluation} 
    \label{flowchart}
\end{figure*}

\subsection{OVERVIEW} \label{overview}

Fig. \ref{flowchart} illustrates the different components of our system. 
Our key insight is to decouple the processes governing tweet, retweet and like activities. This helps to study the effect of different types of activities in news diffusion, and learn an \textit{approximate} model more efficiently than full joint learning. 
%
We map the excitation events to states in a Markov Decision Process (MDP), and learn how to incentivize users 
to increase the \textit{intensity function} for true news diffusion. 
%
To further increase the efficiency, we cluster similar users together so that we can learn a smaller {\em cluster-level} policy, from which we can easily derive individual actions for the members of each cluster. 

Specifically, from the training data we learn a set of Multivariate Hawkes Processes (MHP), one for each type of activity event, and also cluster the users. We use the MHP models to simulate additional training data for learning the policy. While learning the policy function $\pi_C$, we compute the state features for each cluster (based on its current members) and compute the actions for clusters given the current policy, and use those to derive user-level actions. Then we calculate expected reward, and use this to further optimize the policy and update the cluster memberships.



To evaluate the estimated policy $\hat{\pi}_C$, we simulate data again from the MHPs. 
Using the final clusters $C_K$, we obtain actions from the policy to add to the MHP intensity functions and generate evaluation data to assess empirical reward. We also assess the effectiveness of the users selected by our model to promote true news by measuring their number of retweets in held out training data. Each component is described in more detail next.

\vspace{-2mm}
\subsection{ACTIVITY PROCESSES} \label{processes}
\vspace{-1mm}

As in \cite{farajtabar2017fake, goindanisocial}, we use $N$-dimensional MHPs \cite{hawkes1971spectra} to model user activities and simulate the environment dynamics.
%
Let $\lambda_{h, z, i}$ be the intensity function governing the sharing rate of user $i$, where $h$ = $F$ or $T$, and $z$ = $\mathscr{T}$ or $\mathscr{R}$: 

\begin{small}
\begin{equation} \label{MHP_intensity}
\lambda_{h, z, i}(t) = \mu_{h, z, i}  + \sum\nolimits_{j=1}^N  \int_{0}^{t}  \Phi_{z, ji} \text{ } (\omega_{h,z} e^{-\omega_{h,z} t}) \text{ } d\mathcal{N}_j(s, h, z) \nonumber
\end{equation}
\end{small}

\noindent where, the integral is over time, and $s$ is used as placeholder for limits \{0,t\}. 
$\mu_{h, z, i}$ is the base exogenous intensity of user $i$, $\Phi_{z, ji}$ is a kernel adjacency matrix estimated from the training data, and 
$\omega_{h,z} e^{-\omega_{h,z}t}$ is exponential Hawkes kernel. 
For \textbf{Tweet MHP} and \textbf{Like MHP}, we use MHP models proposed in \cite{goindanisocial} to obtain $\mathcal{N}_i(t,h,\mathscr{T})$ and $W_i(t)$.
For the \textbf{Retweet MHP}, we need to estimate the (asymmetric) influence between all pair of users for $\Phi_{\mathscr{R}, ji}$ 
to capture reciprocity.
This naively requires ${\sim}N^2$ parameters, but we use a low-rank approximation of the kernel matrix, proposed in 
\cite{lemonnier2017multivariate}, to improve efficiency. 
%
Our policy will learn actions that correspond to \textit{interventions} to increase sharing of true news events ($h$= $T$) only.
Let $a_{z, k, i}$ be a constant intervention action for user $i$ at stage $k$ (time $t \in [\tau_k, \tau_{k+1})$), added to the her base intensity. 

\begin{equation} \label{general_equation}
\begin{aligned}
\hspace{-2mm}
    \lambda_{T, z, i}(t) &= \mu_{T, z, i} + a_{z, k, i} 
    \\ &+ \sum\limits_{j=1}^N  \int_{0}^{t}  \Phi_{z, ji} \text{ } (\omega_{T,z} e^{-\omega_{T,z} t}) \text{ } d\mathcal{N}_j(s, T, z)
\end{aligned} \nonumber
\end{equation}
\noindent To simulate the event data, we interleave the MHPs to generate tweet events, then retweet events, and then like events.

\subsection{DYNAMIC CLUSTER-BASED POLICY} \label{dynamic_cluster_based_policy}

%
%
Our goal is to learn a policy $\pi$ that maps the state representation (over $N$ users) to intervention actions for tweet and retweet intensities ($N +N^2$ actions). To lower the computational cost, our key insight is to utilize agent correlations to reduce the size of the MDP. 
Specifically, we propose to cluster users into $\mathcal{C}$ clusters, so that we can learn a policy $\pi_C$ that maps the state representation of $\mathcal{C}$ clusters to $\mathcal{C} +\mathcal{C}^2$ actions, where $\mathcal{C} \ll N$. We 
then develop
a method to derive user-level actions from the cluster-level actions. 
%
%
Let $c_{k,i}$ be the cluster of user $i$ at stage $k$, where $c_{k,i}\!=\! m \text{ } (m \in [1, \mathcal{C}]$), and $C_k \!=\! \{c_{k,i}\}_{i=1}^N$ is the set of all cluster assignments at stage $k$.
We define the cluster membership matrix, $\boldsymbol{\mathcal{M}}_k \in \{0, 1\}^{N \times \mathcal{C}}$, at stage $k$, such that $\mathcal{M}_{k, i, m} \!=\! 1$ if $m \!=\! c_{k,i}$, and 0 otherwise. 

Algorithm \ref{cluster_based_policy} outlines our approach to learning a cluster-based policy ${\pi}_C$ parameterized by $\boldsymbol{\theta}$, 
given simulated training data ($\{\textbf{s}_{U,k}\}_{k=1}^{K}$), initial cluster memberships ($\boldsymbol{\mathcal{M}}_1$), and user features ($\textbf{X}_1$) as input, with hyperparameters $\gamma, \eta_\theta, \eta_\phi, \delta$. 
We first compute the state features of the clusters by averaging the state features of their associated members. Then, given the state features per cluster, we apply the current policy to obtain cluster-level actions $\textbf{a}_{C,z,k}$. 
Next, we compute the cluster centroids and derive user-level actions $\textbf{a}_{U,z,k}$, based on the distance of the user to the centroids (see Alg.~\ref{GetUserInterventions}). 
Then we compute the expected reward based on the user-level actions (Alg.~\ref{GetExpectedReward}), and calculate user {\em payoff} and {\em contribution} features $\textbf{X}$ (described later) to recluster users into $\mathcal{C}$ clusters 
(see Alg.~\ref{UpdateClusters}). Finally, we update the policy parameters  $\theta$ by first computing the objective (see Alg.~\ref{GetTotalObjective}) based on the expected reward, and then using stochastic gradient descent with learning rates $\eta_\theta$ and $\eta_\phi$.
The algorithm repeats until convergence and returns the final policy parameters and cluster memberships. 
We describe each component next. 

\begin{algorithm}[!t]
\caption{Dynamic Cluster-based Policy Learning and Optimization} 
\begin{algorithmic}[1]
\State \textbf{Input: $\{\textbf{s}_{U,k}\}_{k=1}^{K}, \boldsymbol{\mathcal{M}}_1, \textbf{X}_1, \gamma, \eta_\theta, \eta_\phi, \delta$}
\Repeat
    \For{$k = 1, ..., K$}
        \State  {\small{/* Compute cluster state features */}}
        \State $\textbf{s}_{C,k,m} \!\!=\!\! \sum\limits_{i=1}^N \mathcal{M}_{k,i, m} \textbf{s}_{U,k,i}/\sum\limits_{i=1}^N \mathcal{M}_{k, i, m}$, $\forall m$
        \State {\small{/* Get cluster-level actions */}}
        \State $\textbf{a}_{C, z, k} = \boldsymbol{\pi}_C(\textbf{s}_{C,k};\boldsymbol{\theta})$ 
        \State  {\small{/* Compute cluster centroids */}}
        \State $\textbf{Y}_{k, m} = {\sum\limits_{i=1}^N \mathcal{M}_{k, i, m} \textbf{X}_{k,i}}/{\sum\limits_{i=1}^N \mathcal{M}_{k,i,m}}$, $\forall m$
        \State {\small{/* Obtain user-level actions with Alg. \ref{GetUserInterventions} ($\pi_U$) */}}
        \State $\textbf{a}_{U, z, k} = \pi_U(\textbf{a}_{C, z, k}, \textbf{X}_k, \textbf{Y}_k)$
        \State $r_k$ = {\small{GetExpectedReward}}($\textbf{s}_{U,k}, \textbf{a}_{U, z, k}, \boldsymbol{\phi}, \gamma$)
        \State {\small{Obtain}} $\textbf{X}_{k+1}$ {\small{using}} $\textbf{a}_{U, z, k}, \textbf{a}_{U, z, k-1}$ (Eq. \ref{pf_past_feature}, \ref{pf_user_expected_contribution_feature})
        \State $\boldsymbol{\mathcal{M}}_{k+1}$ = {\small{UpdateClusters}}($\textbf{Y}_{k}, \boldsymbol{\mathcal{M}}_k, \textbf{X}_{k+1}, \delta$) 
    \EndFor
    \State {\small{/* Learn Policy and update parameters */}}
    \State $ J_\theta, J_\phi$ = {\small{GetTotalObjective}}($\{r_k\}_{k=1}^K, \{\textbf{s}_{U,k}\}_{k=1}^K, \boldsymbol{\phi}, \gamma$)
    \State $\boldsymbol{\theta} = \boldsymbol{\theta} + \eta_\theta \boldsymbol{\nabla_\theta} J_\theta$, \: $\boldsymbol{\phi} = \boldsymbol{\phi} + \eta_\phi \boldsymbol{\nabla_\phi} J_\phi$
\Until{$|\Delta \boldsymbol{\theta}| < \delta$} {\small{/* Convergence */}}
\State $\boldsymbol{\theta}^* = \boldsymbol{\theta}, \boldsymbol{\mathcal{M}}^{*} = \boldsymbol{\mathcal{M}}_{K}$
\\\Return $\boldsymbol{\mathcal{M}}^{*}, \boldsymbol{\theta}^*$
\end{algorithmic}
\label{cluster_based_policy}
\end{algorithm}

\subsubsection{State Features} \label{state_features}

%
We represent the network state $\textbf{s}_k \in \mathbb{R}^{dN}$ 
at stage $k$ as the number of events for $d$ different network activities (i.e. interactions) in the previous stage (e.g. in \cite{qin2015auxiliary}).
Specifically, $s_{k,i,j}$ is the number of events for the $j^{th} (j \!\in\! [1,d])$ activity that user $i \!\in\! [1,N]$ has performed in stage $k\!-\!1$.
%
%
Let $n_{k, i}(h, z) = \mathcal{N}_i(\tau_{k}, h, z) \!-\! \mathcal{N}_i(\tau_{k-1}, h, z)$ represent the number of times user $i$ shares news in stage $k\!-\!1$, and $w_{k, i} = W_i(\tau_{k}) \!-\! W_i(\tau_{k-1})$ represent the number of likes received by user $i$. 
%
%
Let $\textbf{s}_{U,k,i} = (n_{k, i}(T, \mathscr{T}), \break n_{k, i}(F, \mathscr{T}), n_{k, i}(T, \mathscr{R}), n_{k, i}(F, \mathscr{R}), w_{k, i})$ be the state feature for user $i$ that is input to Alg. \ref{cluster_based_policy}. 
We compute the state features for cluster $m$, at stage $k$, as the mean of the state features of its members 
(Line 5, Alg. \ref{cluster_based_policy}). 
Thus, there are $d=5$ network activities corresponding to
tweets (T/F), retweets (T/F), and likes, and
the dimensionality of the state representation 
is $5N$, which will be reduced to $5\mathcal{C}$ 
once we cluster users.

\subsubsection{Reward}

The number of exposures by time $t$ is given by $\sum\nolimits_{i=1}^N \textbf{G}_{i,.} \cdot {\mathcal{N}}_i(t, h, z)$, which 
in stage $k$ is obtained as $\sum\nolimits_{i=1}^N [\textbf{G}_{i,.} \cdot {\mathcal{N}}_i(\tau_{k+1}, h, z) - \textbf{G}_{i,.} \cdot {\mathcal{N}}_i(\tau_{k}, h, z)]$, i.e. 
$\sum\nolimits_{i=1}^N \textbf{G}_{i,.} \cdot {n}_{k,i}(c,z)$. 
Thus, the reward is:

\begin{equation}\label{reward_eq}
\begin{aligned}
    R_z(\textbf{s}_{U,k}) 
    = \frac{1}{N} (\textbf{n}_k(T, z))^\top \textbf{G}^\top \textbf{G} \: \textbf{n}_k(F, z)
\end{aligned}
\end{equation}

\noindent
%
%

%
%

\subsubsection{Objective} \label{objective}

Our goal is to learn a cluster-based policy $\pi_C$ to determine the interventions to be applied to users, at each stage, for true news diffusion process such that the total expected discounted reward for all stages, 
$J = \sum\limits_{k=1}^{K} \gamma^{k} \mathbb{E}[R(\textbf{s}_{U,k}, \textbf{a}_{U, \mathscr{T}, k}, \textbf{a}_{U, \mathscr{R}, k})]$ is maximized, where $\gamma \!\in\! (0,1]$ is the discount rate. 
%
We map the interventions to actions in MDP. 
We impose a budget constraint on the total amount of intervention that can be applied to all users i.e. $||\textbf{a}_{U,z,k}||_1 = B_{z,k}$, where $B_{z,k}$ is the total budget at stage $k$ 
(Line 8, Alg. \ref{GetUserInterventions}).
%
Using Eq. \ref{reward_eq}, we can write,

\begin{equation} \label{expected_reward_eq}
\begin{aligned} 
    \mathbb{E}[R_z(\textbf{s}_{U,k}, \textbf{a}_{U, z, k})] = 
     \frac{1}{N} \mathbb{E}[\textbf{n}_k(T, z)]^\top \: \textbf{G}^\top \textbf{G} \: \: \mathbb{E}[\textbf{n}_k(F, z)] \nonumber
\end{aligned}
\end{equation}

\noindent We assume that the diffusion of fake and true news is independent, and thus, decompose the expected reward. 
See \cite{farajtabar2017fake} for more details on 
$\mathbb{E}[\textbf{n}_k(h, z)]$.
The cumulative expected reward due to tweets and retweets is, $\mathbb{E}[R(\textbf{s}_{U,k}, \textbf{a}_{U, \mathscr{T},k}, \textbf{a}_{U, \mathscr{R},k})] =   \mathbb{E}[R_{\mathscr{T}}(\textbf{s}_{U,k}, \textbf{a}_{U, \mathscr{T}, k})] 
    + \mathbb{E}[R_{\mathscr{R}}(\textbf{s}_{U,k}, \textbf{a}_{U, \mathscr{R}, k})]$.

\subsubsection{Clustering Features} \label{clustering_features}

We design clustering features based on 
{\em Difference Reward} (DR) 
that use a user's contribution to \textit{shape} the reward signal and reduce noise in policy estimates. 
%
We define \textit{payoff} features that indicate how responsive a user is to the policy applied in the past, by measuring the change in a user's expected reward 
after applying policy in stage $k-1$.
%
%

\begin{equation}
\begin{aligned}
    p_{z, k,i} = \mathbb{E}_i[R_z(\textbf{s}_{U,k-1}, \textbf{a}_{U, z, k-1})] - \mathbb{E}_i[R_z(\textbf{s}_{U,k-2}, \textbf{a}_{U, z, k-2})]
\end{aligned} \label{pf_past_feature}
\end{equation}

\noindent
We define \textit{contribution} features $q_{z, k,i}$ 
to measure user $i$'s contribution in the expected reward, given the actions of other users. 
Specifically, we calculate 
the difference in the total expected reward obtained on providing incentive to the user, to when no incentive is provided to the user. To compute the latter, we set the incentive for user $i$ as $0$. 
%
%

\begin{align}
    q_{z, k, i} &= \mathbb{E}_i[R_z(\textbf{s}_{U,k-1}, \textbf{a}_{U, z, k-1})]  \label{pf_user_expected_contribution_feature}
    \\ &- \mathbb{E}_i[R_z(\textbf{s}_{U,k-1}, (\{{a}_{U, z, k, j}\}_{j=1, j \neq i}^N; a_{U, z, k-1, i} = 0))] \nonumber
\end{align} 

\noindent 
Thus, the complete set of features used for clustering users at the start of stage $k$ is 
$\textbf{X}_{k,i} = (p_{\mathscr{T},  k,i}, p_{\mathscr{R}, k,i}, \break  q_{\mathscr{T}, k,i}, q_{\mathscr{R}, k,i}), i \in [1, N]$. The similarity between two users $i$ and $j$ is based on the Euclidean distance between their respective feature vectors $\textbf{X}_{k,i}$ and $\textbf{X}_{k,j}$. 
%
%
%
We do not know the policy estimates apriori for the first stage, and 
obtain initial clusters $C_1$ and membership $\boldsymbol{\mathcal{M}}_1$, using K-means++, 
based on empirical rewards computed from training data. 
%
This captures the \textit{natural policy}  
or intrinsic behavior
of users to spread news, without external incentives. 
We incorporate the DR signals as input to the policy function approximator (via clustering features), rather than using them as explicit shaped reward signal that is different for each agent. 
This helps to 
avoid learning a separate model for each user 
as in the standard DR shaping techniques (e.g. \cite{devlin2014potential}).

\begin{algorithm}[!t]
\caption{GetUserInterventions ($\pi_U$)}
\begin{algorithmic}[1]
\State \textbf{Input: $\textbf{a}_{C, z, k}, \textbf{X}_k, \textbf{Y}_k$} 
    \State {\small{Let $c_{k,i}$ be the cluster to which user $i$ belongs to, at stage $k$.}} 
    \\ {\small{/* Incentive per user for Tweet MHP */}}
    \State $\Tilde{a}_{U, \mathscr{T}, k, i} = a_{C, \mathscr{T}, k, {c_{k,i}}} ||\textbf{X}_{k,i} - \textbf{Y}_{k, {{c_{k,i}}}}||_2$, $\forall i \in [1,N] $
    \State {$\alpha_{U, \mathscr{R}, k, i, j} \!=\! a_{C, \mathscr{R}, k, {{c_{k,i}}, {c_{k,j}}}}  ||\textbf{X}_{k,i} \!\!-\! \textbf{Y}_{k, {{c_{k,i}}}}||_2  ||\textbf{X}_{k,j} \!\!-\! \textbf{Y}_{k, {{c_{k,j}}}}||_2$,} {$\forall i,j \in [1,N] $} {\small{/* Incentive, per pair of users for retweets */}}
    \\{\small{/* Weighted average to get incentive per user for Retweet MHP */}} 
    \State $\Tilde{a}_{U, \mathscr{R}, k, i} =  \frac{1}{N} \sum\limits_{j=1}^N \alpha_{U, \mathscr{R}, k, i, j} \Phi_{\mathscr{R}, ji}$, $\forall i \in [1,N] $ 
    \State $\textbf{a}_{U, z, k} = \frac{\boldsymbol{\Tilde{a}}_{U, z, k}}{||\boldsymbol{\Tilde{a}}_{U, z, k}||_1} \times B_{z,k}$  {\small{/* Budget Constraint */}}
\\\Return $\textbf{a}_{U, z, k}$ 
\end{algorithmic}
\label{GetUserInterventions}
\end{algorithm}

\subsubsection{Learning Cluster-based Policy} \label{policy_learning}




%
Given cluster-level state features $\textbf{s}_{C,k}$, our goal is to learn a cluster-based policy 
parameterized by $\theta$, i.e., $\textbf{a}_{C, \mathscr{T}, k}, \textbf{a}_{C, \mathscr{R}, k} \!=\! \pi_C(\textbf{s}_{C,k}; \boldsymbol{\theta})$.
$\{a_{C, \mathscr{T}, k, m}\}_{m=1}^\mathcal{C}$ is learned for each cluster $C_m$ and corresponds to the incentive for increasing the tweet activities of its members, and $\{a_{C, \mathscr{R}, k, m, {m^\prime}}\}_{m,m^\prime = 1}^\mathcal{C}$ is learned for each \textit{pair} of clusters $(m, {m^\prime})$ and indicates how much to incentivize a user of cluster $m$ to retweet the posts of a user in cluster ${m^\prime}$. 

The value of the network state $\textbf{s}_{U,k}$ 
is the total expected reward when in the given state following policies $\pi_C, \pi_U$, i.e. $V(\textbf{s}_{U,k}) = \mathbb{E}[\sum\limits_{j=k}^K  \gamma^j R_j|(\textbf{s}_{U,k}, \pi_C, \pi_U)]$. 
Since it is computationally expensive to compute $V(\textbf{s}_{U,k})$ from future rewards for every possible policy, we approximate the value as a function of the state parameterized by weights $\boldsymbol{\phi}$, i.e. $V(\textbf{s}_{U,k}) = f(\textbf{s}_{U,k}; \boldsymbol{\phi})$, as in \cite{kurutach2018model}. 
Policy gradient methods are more effective in high dimensional spaces, and can learn continuous policies. Thus, we use 
advantage actor-critic algorithm (e.g. \cite{schulman2015high_A2C}). 

Let there be $K$ stages in the Simulated Training Data (STD) (Fig. \ref{flowchart}). Given state features at the beginning of stage $k$ (i.e. at time $\tau_k$), we learn policy function $\pi_C$ to obtain actions to be applied during stage $k$ (i.e. time-interval $[\tau_{k}, \tau_{k+1})$), using a multi-layer feed-forward neural network (NN) (see \cite{goindanisocial} for details on NN).
%
%
%
%
%
We find intervention actions for the clusters, $\textbf{a}_{C, \mathscr{T}, k}, \textbf{a}_{C, \mathscr{R}, k} = \pi_C(\textbf{s}_{C,k})$, corresponding to tweet and retweet intensities, respectively, as output of the NN 
(line 7 of Alg. \ref{cluster_based_policy}). 
Then, in line 11, we obtain actions for users $\textbf{a}_{U, z, k}$ based on their \textit{variability} from their cluster, using Alg. \ref{GetUserInterventions}.
%
%
Alg. \ref{GetUserInterventions} computes the actions for the users by weighting the cluster actions by their distance to the centroid in lines 4 and 5. 
This helps to capture both similarity and variability in user behaviors.
%
%
%
Since retweets involve an interaction between pair of users, 
we consider incentives $\alpha_{U, \mathscr{R}, k, i,j}$ for each pair of users, where $\alpha_{U, \mathscr{R}, k, i,j}$ is the amount of incentive provided to user $i$ to retweet user $j$'s posts. However, due to high computational cost, we do not want to model an $N^2$-dimensional MHP for retweet activities. So instead, to reduce it to an $N-$dimensional MHP, we compute the weighted average of incentive actions 
in line 7.
We normalize user actions based on our budget constraint in line 8. 
%
%
Then, we compute the expected reward using these actions, 
as described in Alg. \ref{GetExpectedReward},
%
%
\begin{algorithm}[!t]
\caption{GetExpectedReward}
\begin{algorithmic}[1]
\State \textbf{Input: $\textbf{s}_{U,k}, \textbf{a}_{U, z, k}, \boldsymbol{\phi}, \gamma$} 
        \State Compute $\mathbb{E}[R_{k}(\textbf{s}_{U,k}, \textbf{a}_{U, z, k})]$ (Sec. \ref{dynamic_cluster_based_policy}, `Objective') 
        \State $\textbf{s}_{U, k^\prime} = (\mathbb{E}[\textbf{n}_k(h, \mathscr{T})], \mathbb{E}[\textbf{n}_k(h, \mathscr{R})], \mathbb{E}[\textbf{w}_k])$ 
        \State $V(\textbf{s}_{U,k^\prime}) = f(\textbf{s}_{U,k^\prime}; \boldsymbol{\phi})$
        {\small{\State $r_k = \mathbb{E}[R_{k}(\textbf{s}_{U,k}, \textbf{a}_{U, \mathscr{T}, k}, \textbf{a}_{U, \mathscr{R}, k})] + \gamma V(\textbf{s}_{U, k^\prime})$}
        }
\\\Return $r_k$
\end{algorithmic}
\label{GetExpectedReward}
\end{algorithm}
\begin{algorithm}[!t]
\caption{GetTotalObjective}
\begin{algorithmic}[1]
\State \textbf{Input: $\{r_k\}_{k=1}^K, \{\textbf{s}_{U,k}\}_{k=1}^K, \boldsymbol{\phi}, \gamma$} 
    \State $L_\theta = 0$, \: $L_\phi = 0$
    \For{$k = 1, ..., K$}
        \State $V(\textbf{s}_{U,k}) = f(\textbf{s}_{U,k}; \boldsymbol{\phi})$ {\small{/* Value Function */}}
        \State Let $D_k = \sum \limits_{j=k}^{K} \gamma^k r_k$ {\small{/* Total Discounted Reward */}}
        \State $B_k = D_k - V(\textbf{s}_{U,k})$ {\small{/* Advantage Function */}}
        \State $L_\theta = L_\theta +  B_k$; \hspace{2mm} $L_\phi = L_\phi +  ||V(\textbf{s}_{U,k}) -  D_k||_2 $
    \EndFor
    \State $ J_\theta = L_\theta $, \: $J_\phi = - L_\phi $
\\\Return $J_\theta, J_\phi$
\end{algorithmic}
\label{GetTotalObjective}
\end{algorithm}
and the total objective for optimizing policy $\pi_C$ in Alg. \ref{GetTotalObjective}.
\begin{algorithm}[!t]
\caption{UpdateClusters}
\begin{algorithmic}[1]
\State \textbf{Input: $\textbf{Y}_{k}, \boldsymbol{\mathcal{M}}_k, \boldsymbol{X}_{k+1}, \delta$} 

    \State Let $\boldsymbol{\ddot{Y}}_{k+1} = \textbf{Y}_{k}$, $\boldsymbol{\mathcal{M}}_{k+1} = \boldsymbol{\mathcal{M}}_k$
    \Repeat
        \State Let $\boldsymbol{\dot{y}} = \boldsymbol{\ddot{Y}}_{k+1}$ {\small{/* Centroids in previous iteration */}}
        \State $\boldsymbol{\dot{Y}}_{k+1, m} =  (\sum\limits_{i=1}^N \mathcal{M}_{k+1,i,m} \textbf{X}_{k+1, i}) / (\sum\limits_{i=1}^N \mathcal{M}_{k+1,i,m})$ 
        \State $\boldsymbol{\ddot{Y}}_{k+1, m} = \varepsilon_1 \boldsymbol{\dot{Y}}_{k+1, m} + \varepsilon_2 \textbf{Y}_{k, m}$
        \State $m_i = \arg \max \limits_{m = 1}^\mathcal{C} || \textbf{X}_{k+1,i} - \boldsymbol{\ddot{Y}}_{k+1, m}||^2$ 
        \State $\mathcal{M}_{k+1, i, m_i} = 1$, and $\forall m \neq m_i, \mathcal{M}_{k+1,i,m} = 0$
    \Until{$||\boldsymbol{\dot{y}} - \boldsymbol{\ddot{Y}}_{k+1}||_2 < \delta$} {\small{/* Convergence */}}
\\\Return $\boldsymbol{\mathcal{M}}_{k+1}$
\end{algorithmic}
\label{UpdateClusters}
\end{algorithm}

\subsubsection{Update Clusters}

Using actions $\textbf{a}_{U, z,k}$ learnt for stage $k$, we calculate the clustering features (Eq. \ref{pf_past_feature}-\ref{pf_user_expected_contribution_feature}) for the next stage, $\textbf{X}_{k+1}$. 
Due to application of the policy, $\textbf{X}_{k+1,i}$ is different from $\textbf{X}_{k,i}$. Thus, we need to re-compute the centroids and cluster memberships.
%
%
Additionally, we want the clusters to be aligned across different stages, so that
the policies can be optimized using the neural network, for different clusters across multiple epochs.
To achieve this, we define \textit{weighted centroids} that include the effect of centroids in the previous stage. 
This helps to ensure that the centroids do not shift much and thus, we can
align clusters in stage $k+1$ with those in stage $k$. 
Alg. \ref{UpdateClusters} shows the steps to update clusters.
%
%
%
For clustering at stage $k+1$, we begin with the memberships from stage $k$ (line 2). This helps in faster convergence of clusters. 
Using $\textbf{X}_{k+1}$, we obtain the updated centroids $\boldsymbol{\dot{Y}}_{k+1, m} \forall m \in [1, \mathcal{C}]$ for stage $k+1$ (line 5). We define $\mathcal{C}$ \textit{weighted centroids} $\boldsymbol{\ddot{Y}}_{k+1, m} = \varepsilon_1 \boldsymbol{\dot{Y}}_{k+1, m} + \varepsilon \textbf{Y}_{k, m} \forall m \in [1, \mathcal{C}]$, (line 6), $\varepsilon_1 + \varepsilon_2 = 1$. $\varepsilon_1$, $\varepsilon_2$ indicate the importance assigned to the centroids from the previous stage and the current stage, respectively.
After updating the centroids, we update the membership matrix and repeat until  convergence. 
%
Additionally, since the change in policy estimates  across epochs reduces as optimization gets closer to convergence, 
the clustering features (which are dependent on these estimates) do not change much for the same stages across such epochs and we can start to reuse the learned clusters. 
%
Similar to simulated annealing, we only update the cluster assignments every $\eta_e \in \mathbb{Z}^{+}$ epochs gradually increasing $\eta_e$ as the epoch number increases, which helps speed up convergence of policy learning.
%




\subsection{POLICY EVALUATION}

Since the Simulated Evaluation Data (SED) 
is conditioned on STD, we use the clusters converge at the end of the policy learning 
to evaluate the estimated policy. 
This helps to avoid re-clustering at each stage of evaluation, and reduce running time for online evaluation. 
%
First, we find intervention actions 
for the network state obtained from events in SED. 
%
Then, we generate events by simulating MHPs after adding interventions to the base intensities for true news diffusion, 
and use those to compute the following evaluation metric.
%

\subsubsection{Evaluation Metric}


Comparing different methods solely based on reward (e.g. \cite{farajtabar2017fake}) does not suffice, and a better model is the one that mitigates more distinct number of users \cite{goindanisocial}.
Thus, instead of considering only the reward as in previous work (e.g. \cite{farajtabar2017fake}), 
we multiply the reward by the the fraction of users exposed to fake news that become exposed to true news, as also used in \cite{goindanisocial}.
This helps to assign more importance to the selection of distinct users over selection of few users with high exposures. Specifically, performance $\mathcal{P}$ is given as $\sum\limits_{k=1}^K R_k \times \frac{|L_{T,k} \cap L_{F,k}|}{|L_{F,k}|}$, where $L_{T,k} = \{i| i \in [1, N], \textbf{n}_{k}(T,z) \cdot \textbf{G}_{.i} > 0\}$ and $L_{F,k} = \{i|i \in [1, N], \textbf{n}_{k}(F,z) \cdot \textbf{G}_{.i} > 0\}$ are the sets of users exposed to true and fake news, respectively, during stage $k$. 

\section{EXPERIMENTS}

We use real-world datasets, Twitter 2016 and Twitter 2015 \cite{ma2017detect_twitter_2016_2015,liu2018early}, with 749 and 2051 users, respectively. 
We consider time-horizon $\mathcal{T} = 40$, divided into 40 stages of $\Delta T = 1$ hour each. 
%
%
%
The Training Data, STD, SED before and after applying interventions, and Held-Out Data (Fig.~\ref{flowchart}), respectively, correspond to time-intervals, $[0, 10)$, $[10, 20)$, $[20, 30)$, and $[30, 40)$.
We use ``tick" python library \cite{2017arXiv170703003B_tick} to perform
simulations of MHPs. 
We consider $O_{z,k} \sim N \cdot \mathcal{U}(0,1)$, and $\gamma = 0.7$, as in \cite{goindanisocial, farajtabar2017fake}. 

\begin{figure}[!t]
\hbox{
\subfloat[Fake News]{\includegraphics[scale = 0.23]{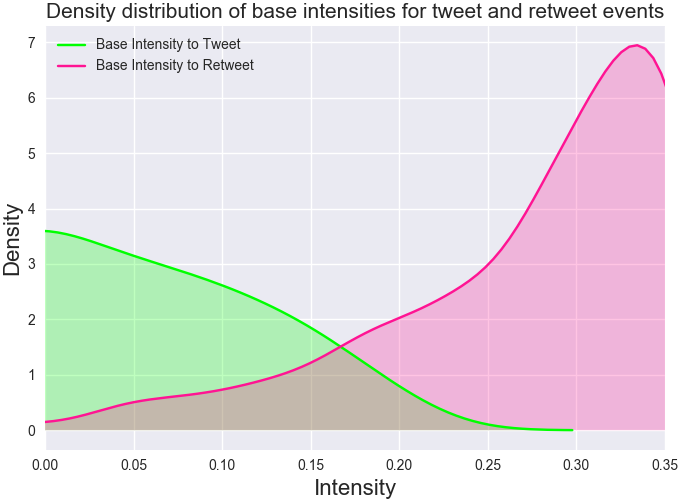}}
\subfloat[True News]{\includegraphics[scale = 0.23]{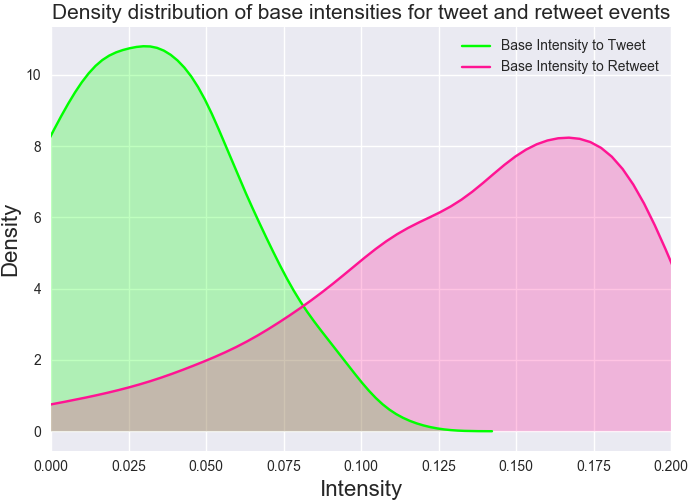}}
}
\caption{Distribution for Tweet and Retweet Intensities}
\label{comparison_base_intensity_tweet_retweet_2015}
\end{figure}

First, we test whether news diffusion patterns via tweet and retweet events are similar in the Twitter data. 
Fig. \ref{comparison_base_intensity_tweet_retweet_2015} shows a comparison of the distribution of respective base intensities across all users. 
We observe differences in the distribution of base intensities, which 
supports our approach
to consider separate network activities (tweet and retweet) to characterize different types of agent interactions. 

%

\subsection{BASELINES}

In our best knowledge, our approach \textbf{\MethodName}, is the first to consider policy learning while dynamically clustering users in MARL for social networks and there are no other alternative dynamic approaches to compare to. We designed a dynamic Rule-based clustering approach, which uses thresholds to (re)group users, to use as a baseline for comparison. But it was always outperformed by \MethodName and since it’s a naive baseline, we did not include it due to space limit. 
%
%
Thus, we compare to the static clustering methods \cite{zhu2018group, el2018personalization}, and the non-clustering methods that have also been used in \cite{zhu2018group, el2018personalization}, as described below.

\noindent \underline{\emph{Non-Clustering Methods}}

\paragraph{NC-1}

No Clustering. All users in same cluster ($\mathcal{C} = 1$). 


\paragraph{NC-N}

MHP-U model in \cite{goindanisocial}: Identical actions learnt for tweet and retweet activities. Each user is in a separate cluster i.e. $\mathcal{C} = N$. We primarily compare to NC-N since 
it outperforms different non-clustering approaches (eg. \cite{farajtabar2017fake}).

\paragraph{NC-TR}

Separate actions learnt for tweet and retweet activities (for true news diffusion), and $\mathcal{C} = N$.

\paragraph{NC-PF}


Same as NC-TR, but with clustering features (Eq. \ref{pf_past_feature}, \ref{pf_user_expected_contribution_feature}) added in the state representation of users, and $\mathcal{C} = N$.

\noindent \underline{\emph{Clustering Based Methods}}

\paragraph{RND}

Randomly assign users to $\mathcal{G}$ static (fixed) groups. 

\paragraph{C-NET}

Clusters obtained using K-Means++ with network features (degree, closeness centrality, clustering coefficient). 

\paragraph{KM-R}

%

Static Clusters are obtained using K-Means++ with empirical reward features, and are not updated dynamically.

\paragraph{KM-S}

\cite{el2018personalization, zhu2018group}.
Static Clusters as in KM-R, but state features are also used for clustering.

The input to the policy function approximator (Sec. \ref{policy_learning}) 
requires input and output of fixed dimensions, and hence, we assume fixed number of clusters $\mathcal{C}$. We use the scores of Bayesian Information Criterion, and Within Cluster Sum of Squared Distance to
select $\mathcal{C}$ for each clustering-based method. 
%
%
Table \ref{num_clusters_all_methods} reports the number of clusters ($\mathcal{C}$) obtained for different baselines, and we observe
that $C \in \{8,9\}$ is similar for different methods. 
%
%

\begin{table}[!b]
\centering
\caption{Number of Clusters ($\mathcal{C}$) for Different Methods}
\label{num_clusters_all_methods}
\scalebox{0.9}
{
\begin{tabular}{cccccc}
    \toprule
    & \textbf{\MethodName} & \textbf{KM-R} & \textbf{KM-S} & \textbf{C-NET} & \textbf{RND} \\
    \midrule
    Twitter 2016 & 9  & 9 & 9 & 8 & 9 \\
    Twitter 2015 & 8  & 8 & 9 & 8 & 9 \\
\bottomrule
\end{tabular}
}
\label{table_results}
\end{table}

\subsection{RESULTS}




%
%
\begin{table}[t!]
\centering
\caption{Relative Performance (Mean $\pm$ Std. Error)}
\label{relative_perf_all_methods}
\scalebox{0.88}
{
\begin{tabular}{c|c|c|c}
    \toprule
  {} &  {} & \textbf{Twitter 2016} & \textbf{Twitter 2015} \\
    \midrule
    \rowcolor{Yellow}
   \textbf{Clustering} & {DCPL}  & \textbf{98.19} $\pm$ 1.52  &  \textbf{95.175} $\pm$ 1.75 \\
   \rowcolor{Yellow}
   {Based} & {KM-R} & 81.28 $\pm$ 1.78  &  73.98 $\pm$ 1.64  \\
   \rowcolor{Yellow}
   {Methods} & {KM-S} & 78.63 $\pm$ 1.82  & 70.37 $\pm$ 1.69\\
   \rowcolor{Yellow}
    & {C-NET} & 64.07 $\pm$ 1.98  &  51.52 $\pm$ 1.72 \\
    \rowcolor{Yellow}
    & {RND} & 55.52 $\pm$ 4.23  &  42.17 $\pm$ 4.94\\
     \cmidrule{1-4}
     \rowcolor{Green}
   \textbf{Non-Clustering} &  NC-PF  & 87.39 $\pm$ 3.46  & 80.73 $\pm$ 3.72   \\
   \rowcolor{Green}
   {Methods} & {NC-TR} & 77.83 $\pm$ 3.37  &  62.76 $\pm$ 3.51 \\
   \rowcolor{Green}
   {} & {NC-N} & 67.04 $\pm$ 3.21  &  56.10 $\pm$ 3.48 \\
    \rowcolor{Green}
    & NC-1 & 58.68 $\pm$ 1.02  & 47.12 $\pm$ 1.10 \\
    \bottomrule
\end{tabular} 
}
\end{table}

Table \ref{relative_perf_all_methods} shows the relative performance of different methods. \MethodName is able to achieve highest reward. Also, NC-TR outperforms NC-N, implying that it is beneficial to decouple the different network activities. The clustering-based methods achieve greater performance than non-clustering methods.


NC-N, NC-TR, NC-PF have larger variance and noisy estimates due to high dimensionality of state/action space. 
NC-1 has high bias as it doesn't consider differences in agent behavior. Clustering based approaches achieve lower variance, 
which indicates that they
overcome sparsity and curse of dimensionality. 
%
%

Moreover, \MethodName that updates cluster assignments dynamically based on the policy applied outperforms those that assume fixed assignments (KM-R, KM-S, C-NET, RND). 
Notably, NC-PF that does not perform clustering 
also outperforms these, 
as it adjusts policy based on dynamic agent behavior. 
Thus, we need features indicative of agents' \textit{payoff} and \textit{contribution}, apart from state features, to get more accurate 
estimates.
%

The performance of KM-R is slightly greater than KM-S, implying that reward based features alone are useful for learning better estimates,
without state features.
Notably, C-NET method that uses network-based features performed poorly. This indicates that not all similarity measures are useful for obtaining \textit{good} clusters, i.e. ones that increase efficiency without reducing effectiveness. These results support the claim that features indicative of expected reward and agent contribution are likely to be more helpful.
%


%
\begin{figure}[!t]
\hbox{
\subfloat[Twitter 2016]{\includegraphics[scale = 0.275]{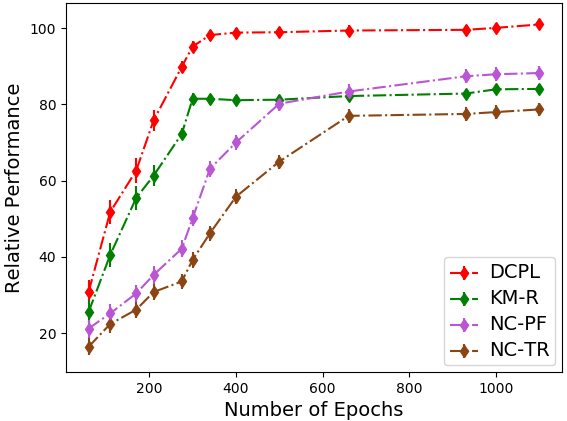}}
\hspace{1ex}
\subfloat[Twitter 2015]{\includegraphics[scale = 0.275]{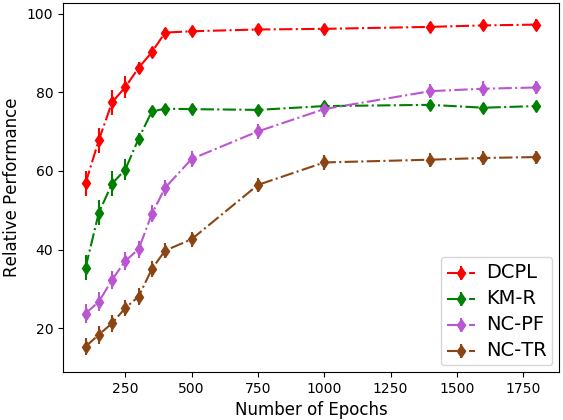}}
}
\caption{Number of Epochs until Convergence}
\label{num_epochs}
\end{figure}
%
%
Fig. \ref{num_epochs} shows a comparison of the time taken until convergence 
by different methods.
\MethodName converges faster than NC-PF and NC-TR, and achieves a greater performance for all epochs. 
%
NC-PF outperforms KM-R and NC-TR, but takes longer to converge, implying that the reward-based features (Eq. \ref{pf_past_feature}, \ref{pf_user_expected_contribution_feature}) 
are useful if included in state representation, however, lead to increased computational cost. 
%
%
In \MethodName, the information about agents' payoff and contribution via clusters, helps in better exploration over the action space without increasing the state space. This helps the model to learn agents' effectiveness early on and converge faster to better policy. 

The above results, based on reward, serve as preliminary proof of concept that providing incentives helps to mitigate the impact of fake news \cite{goindanisocial}. Since we cannot make real-time interventions, we also compare different methods by measuring the impact of nodes selected for intervention, in terms of the people they actually reached in the real Held-Out Data.
Let $S(\tau)$ refer to the the set of users \textit{selected} in SED ($\tau \in [20, 30)$) to spread true news by time $\tau$, according to the model, i.e., $S(\tau) = \{i|(\mathcal{N}_i(\tau, T, z) - \mathcal{N}_i(20, T, z)) > 0\}$, 
and the remaining users are considered \textit{missed} ($M(\tau)$) by the model. We calculate the total number of users who retweeted the posts of users in $S(\tau)$ and $M(\tau)$ between time $[\tau^{'}, \tau^{'} + \Delta)$ where $\tau^{'} \!=\! \tau + g$, and 
$g \!=\! \{0, 2, 5, 8\}$ indicates the gap or number of stages after which we want to measure the impact of users (in the future). We considered different values of $\Delta \in \{1, 2, 3, 4, 5\}$ and Table \ref{comparison_impact_future} reports the average. We see that the impact of selected nodes (S) is greater than that of missed nodes (M) for \MethodName by a large margin. 
\begin{table}[!t]
\centering
\caption{Sum of retweets at $\tau + g$ for users selected at $\tau$}
\scalebox{0.73}
{
\hspace{-6mm}
\begin{tabular}{crrrrrrrr}
    \toprule
    \multirow{2}{*}[-0.5\dimexpr \aboverulesep + \belowrulesep + \cmidrulewidth]{{MODEL}}  & \multicolumn{2}{c}{$\tau^{'}=\tau + 0$}
    & \multicolumn{2}{c}{$\tau^{'}=\tau + 2$} & \multicolumn{2}{c}{$\tau^{'}=\tau + 5$} & \multicolumn{2}{c}{$\tau^{'}=\tau + 8$} \\
    \cmidrule(l){2-3} \cmidrule(l){4-5} \cmidrule(l){6-7} \cmidrule(l){8-9}
    & {S} & {M} & {S} & {M} & {S} & {M} & {S} & {M}  \\
    \midrule
    \rowcolor{Yellow}
DCPL & \textbf{1320.9} & \textbf{439.8} & \textbf{1289.1} & {393.2} & \textbf{912.5} & \textbf{352.4} & \textbf{853.3} & \textbf{208.2} \\ 
\rowcolor{Yellow}
KM-R & 1181.2 & 546.6 & 875.6 & 381.2 & 667.4 & 419.6 & 473.2 & 366.8 \\
\rowcolor{Yellow}
KM-S & 1103.7 & 510.8 & 830.2 & \textbf{311.3} & 660.4 & 415.5 & 492.2 & 356.6 \\
\rowcolor{Yellow}
C-NET & 810.4 & 691.2 & 695.3 & 669.6 & 517.8 & 545.7 & 496.5 & 453.6 \\
\rowcolor{Yellow}
RND & 569.9 & 712.5 & 485.1 & 600.6 & 369.7 & 706.8 & 329.2 & 591.3 \\
\cmidrule{1-9}
\rowcolor{Green}
NC-PF & 1200.4 & 471.4 & 890.7 & 406.2 & 682.6 & 397.6 & 593.2 & 283.4 \\
\rowcolor{Green}
NC-TR & 1077.2 & 522.4 & 798.4 & 425.8 & 652.5 & 450.6 & 400.4 & 274.4 \\
\rowcolor{Green}
NC-N & 1022.5 & 498.8 & 735.1 & 378.3 & 621.8 & 378.3 & 528.3 & 297.6 \\ 
\bottomrule
\end{tabular}
}
\label{comparison_impact_future}
\end{table}

\begin{figure}[!t] 
\hbox{
\subfloat[Twitter 2016]{\includegraphics[scale = 0.265]{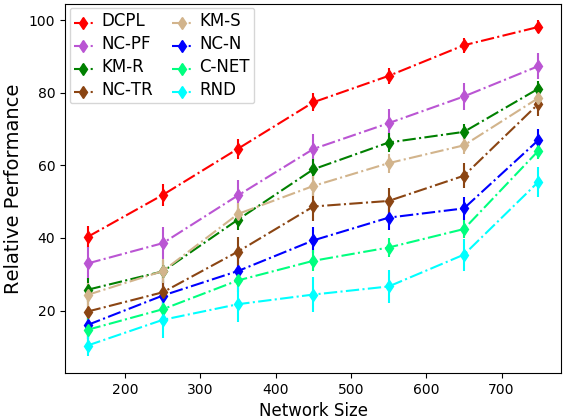}}
\hspace{1ex}
\subfloat[Twitter 2015]{\includegraphics[scale = 0.265]{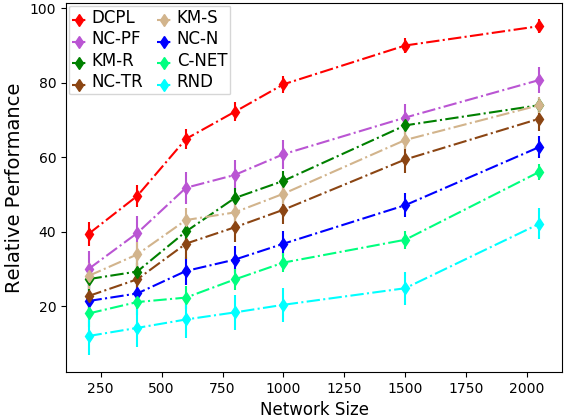}}
}
\caption{Relative Performance vs $N$}
\label{rel_perf_vs_network_size}
\end{figure}

We also explored the effects of network characteristics on performance. 
We down-sampled the datasets to compare the performance of different approaches as a function of network size $N$. Fig. \ref{rel_perf_vs_network_size} shows that the performance of all methods decreases with a decrease in $N$, and our method \MethodName outperforms for all network sizes considered.
%
%

\begin{figure}[!t]
\centering 
    {\includegraphics[width = 4cm]{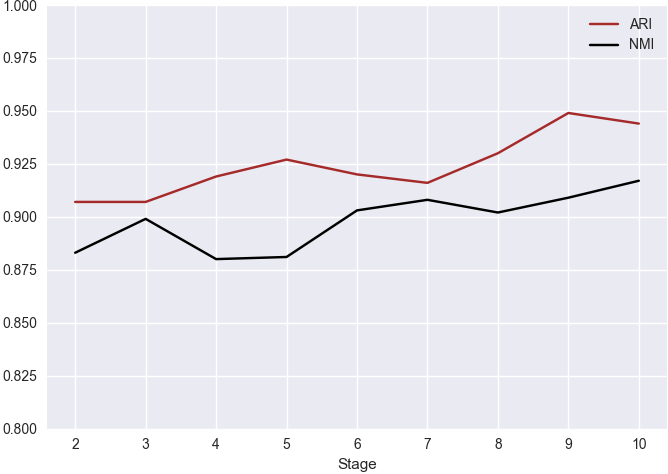}}
    \caption{Cluster Alignment Scores for DCPL}
    \label{cluster_alignment}
\end{figure}

We also conducted additional experiments to assess the learned clusters. 
To evaluate cluster alignment across different stages, in our method \MethodName,
we compare the clusters at stage $k \geq 2$ with those obtained in the previous stage $k-1$. 
Figure \ref{cluster_alignment} shows the Adjusted Rand Index (ARI) and Normalized Mutual Information (NMI) scores. For stage 2, we compare with the initial clusters (Sec. \ref{clustering_features}) . 
%
%
%
We see that both NMI and ARI scores are high, and this presents a proof of concept that clusters of \MethodName are indeed aligned.

\begin{figure}[!b]
\hbox{
\subfloat[Twitter 2016]{\includegraphics[scale = 0.23]{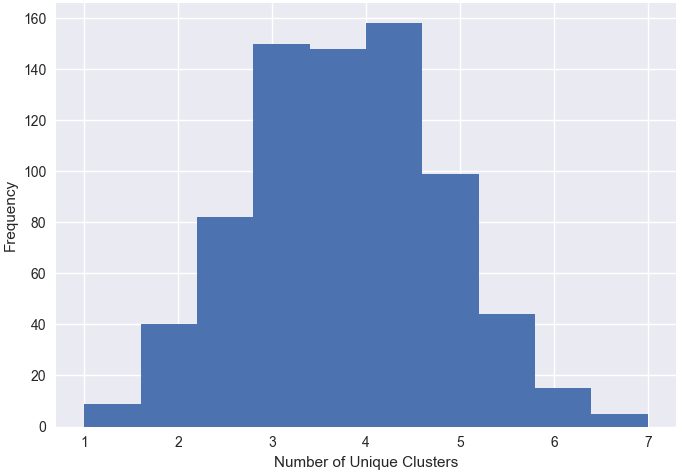}}
\hspace{1ex}
\subfloat[Twitter 2015]{\includegraphics[scale = 0.23]{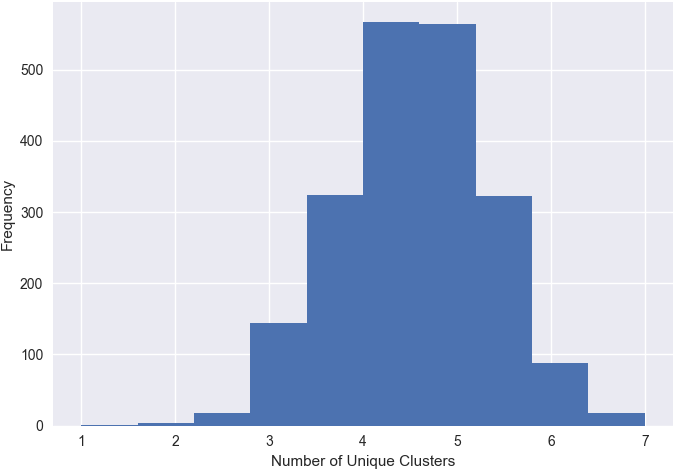}}
}
\caption{Number of unique clusters across users}
\label{num_unique_clusters}
\end{figure}
\begin{figure}[!b]
\hbox{
\subfloat[Twitter 2016]{\includegraphics[scale = 0.23]{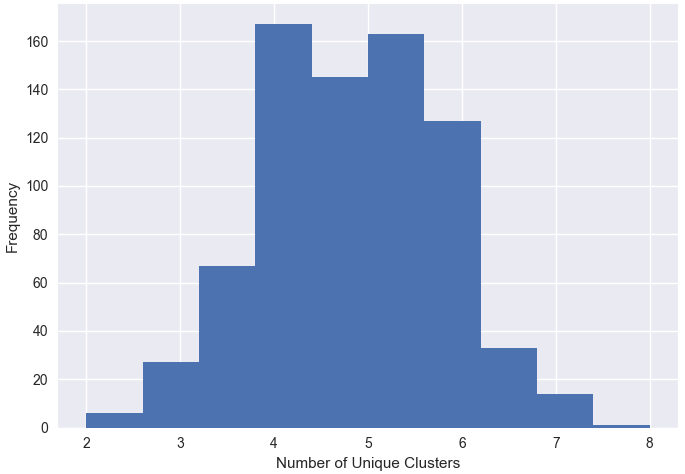}}
\hspace{1ex}
\subfloat[Twitter 2015]{\includegraphics[scale = 0.23]{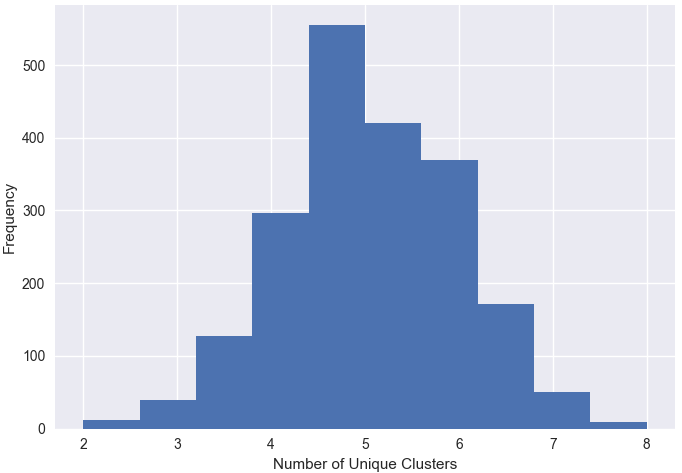}}
}
\caption{Number of cluster transitions across stages}
\label{num_change_clusters}
\end{figure}

We 
studied the movement of users between different clusters across multiple stages. Fig. \ref{num_unique_clusters} and \ref{num_change_clusters}, respectively, show the distribution for the number of unique clusters that a user is in, and the number of times a user changes clusters, across all stages in the learning phase. 
We observe that users generally change clusters and do not return back to the same clusters. 
This shows that there is a change in user behavior (features) due to the policy applied in the past, justifying our approach to 
update cluster assignments dynamically.
%

%


\begin{figure}[!b]
\hbox{
\subfloat[DCPL, KM-R]{\includegraphics[scale = 0.35]{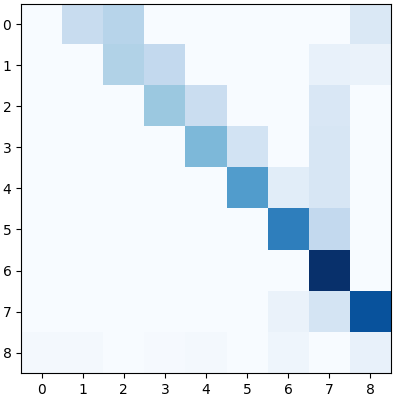}}
\hspace{3ex}
\subfloat[DCPL, KM-S]{\includegraphics[scale = 0.35]{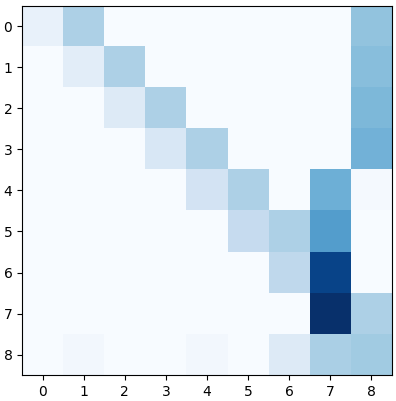}}
}
\hbox{
\subfloat[DCPL, C-NET]{\includegraphics[scale = 0.35]{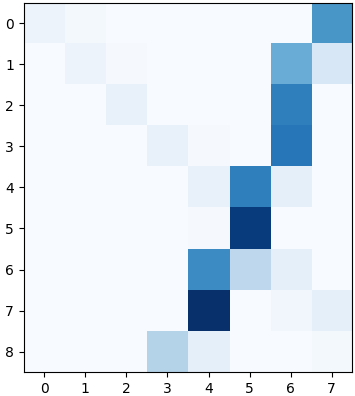}}
\hspace{4ex}
\subfloat[DCPL, RND]{\includegraphics[scale = 0.35]{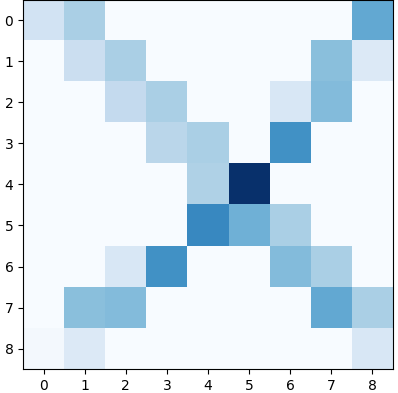}}
}
\caption{Contingency Matrix (DCPL vs other baselines)}
\label{comparison_clusterings}
\end{figure}
%
%
Fig. \ref{comparison_clusterings} shows contingency matrices comparing the clusters obtained by our method \MethodName with those obtained by other baselines. 
%
%
We compare the clusters obtained by different methods, and find that clusters of \MethodName are most similar to those of KM-R, followed by KM-S, C-NET, RND, consistent with the decreasing 
order of these methods, in terms of maximizing reward.

\section{CONCLUSION} \label{conclusion}



This paper outlines a Dynamic Cluster-based Policy Learning approach, DCPL, for social reinforcement learning that considers domains with a large number of agents with complex interactions among them. Our approach helps to address the challenges of high-dimensionality and sparsity by exploiting the social network structure and agent similarities to obtain a compact model that better captures agent dependencies and is much more efficient to solve. 

Specifically, we use latent features to cluster users based on their payoff and contribution, and aggregate the interactions of similar agents. We learn policies for the clusters, and then easily use those to obtain actions for users based on their variability from the cluster. This allows for efficiently learning personalized policies, while still considering all agent dependencies given large number of users. Moreover, the number of effective policies is greatly reduced, thus lowering the computational complexity. 
Clustering adds a discriminative power to our model to differentiate between different types of users (and their activities), which helps in efficient and effective allocation of incentives among users under fixed budget.
Experiments show that dynamic clustering of users 
helps the model to quickly learn better policy estimates, allowing it to outperform other static clustering-based and non-clustering alternatives. 

\section*{Acknowledgements}
We thank Guilherme Gomes for constructive discussions. This research is supported by NSF and AFRL under contract numbers IIS-1618690 and FA8650-18-2-7879. 

\bibliographystyle{abbrv}
{\small \bibliography{arxiv_full}}

\end{document}